\documentclass[letterpaper, 10 pt, conference]{ieeeconf}  

\IEEEoverridecommandlockouts                              

\usepackage{graphicx} 
\usepackage{mathtools}
\usepackage{amsmath} 

\usepackage{textcomp}

\title{\LARGE \bf
 An Efficient and Scalable Collection of Fly-inspired Voting Units for Visual Place Recognition in Changing Environments
}

\author{Bruno Arcanjo$^{1}$, Bruno Ferrarini$^{1}$, Michael Milford$^{2}$, Klaus D. McDonald-Maier$^{1}$ and Shoaib Ehsan$^{1}$
\thanks{*This work was supported by the UK Engineering and Physical Sciences Research Council through grants EP/R02572X/1, EP/P017487/1, and in part by the RICE project funded by the National Centre for Nuclear Robotics Flexible Partnership Fund.}
\thanks{$^{1}$B. Arcanjo, B. Ferrarini, K. D. McDonald-Maier and S. Ehsan are with the School of Computer Science and Electronic Engineering, University of Essex, United Kingdom {\tt\small (email: bq17319@essex.ac.uk; bferra@essex.ac.uk; kdm@essex.ac.uk; sehsan@essex.ac.uk)}}%
\thanks{$^{2}$M. Milford is with the School of Electrical Engineering and Computer Science, Queensland University of Technology, Brisbane, QLD 4000, Australia
        {\tt\small (email: michael.milford@qut.edu.au)}}%
}

\begin{document}

\maketitle
\thispagestyle{empty}
\pagestyle{empty}

\begin{abstract}

State-of-the-art visual place recognition performance is currently being achieved utilizing deep learning based approaches. Despite the recent efforts in designing lightweight convolutional neural network based models, these can still be too expensive for the most hardware restricted robot applications. Low-overhead VPR techniques would not only enable platforms equipped with low-end, cheap hardware but also reduce computation on more powerful systems, allowing these resources to be allocated for other navigation tasks. In this work, our goal is to provide an algorithm of extreme compactness and efficiency while achieving state-of-the-art robustness to appearance changes and small point-of-view variations. Our first contribution is DrosoNet, an exceptionally compact model inspired by the odor processing abilities of the fruit fly, Drosophyla melanogaster. Our second and main contribution is a voting mechanism that leverages multiple small and efficient classifiers to achieve more robust and consistent VPR compared to a single one. We use DrosoNet as the baseline classifier for the voting mechanism and evaluate our models on five benchmark datasets, assessing moderate to extreme appearance changes and small to moderate viewpoint variations. We then compare the proposed algorithms to state-of-the-art methods, both in terms of precision-recall AUC results and computational efficiency.

\end{abstract}

\section{Introduction}
Visual place recognition (VPR) refers to the ability of a computer system to determine if it has previously visited a given place using visual information. Performing highly robust and reliable VPR is a key feature for autonomous robotic navigation as SLAM systems are dependent on loop-closures mechanisms for map correction \cite{ref:vpr-survey}. While the VPR problem is well-defined, it remains an extremely difficult task to perform reliably as there are a range of challenges that must be dealt with. Firstly, a revisited place can look extremely different from when it was first seen and recorded due to a variety of changing conditions: seasonal changes \cite{ref:season_changes}, different viewpoints \cite{ref:pov_changes}, illumination levels \cite{ref:illu_changes}, dynamic elements \cite{ref:dyna_changes} or any combination of these factors. It is also possible for different places to appear identical, especially within the same environment, an error known as perceptual aliasing. 

Initially used for difference computer vision tasks, convolution neural network (CNN) based models have been made their way into the VPR field over recent years, achieving impressive performance on a variety of datasets \cite{ref:VPR_CNNs}. However, real-time VPR CNN approaches often rely on powerful graphic processing units (GPUs) and large amounts of memory, making them unsuitable for extremely hardware restricted applications \cite{ref:res_hardware_1}. Mobile robotics with resource-constrained hardware are common and these systems cannot afford to run such computationally expensive algorithms \cite{ref:res_hardware_1}, \cite{ref:res_hardware_2}. VPR techniques which manage to keep memory usage and computational complexity to a minimum, without compromising performance, are key to enable platforms equipped with low-end hardware. Furthermore, low-overhead VPR algorithms would also benefit systems that are able to run expensive models, freeing resources that can be allocated to other essential functionalities of a robot's navigation. This is the motivation for the recent development of several low-overhead alternatives \cite{ref:flynetcann}, \cite{ref:squeezenet}, \cite{ref:mobilenets} and in this work we continue to add on to this literature.

In this paper, we start by presenting a lightweight biological-inspired algorithm dubbed DrosoNet, designed after the brain of drosophila melanogaster \cite{ref:fruitfly} and its ability to recognize odors by encoding complex patterns in a small representation tag. DrosoNet features a low model size of 190KiBs and an inference time of around 1 millisecond, making it both extremely compact and computationally efficient. However, performance is compromised when compared to state-of-the-art models and while robustness to extreme appearance changes and moderate viewpoint shifts is promising, most applications require more reliable VPR.

Our solution to DrosoNet's compromised performance was to utilize multiple of these small models in conjunction, made possible by the low memory size algorithm features. Furthermore, there is intrinsic randomness to DrosoNet's training process, both in the model's initialization and in it's fully connected layer, which results in variation of its predictions in deployment. The key observation is that one DrosoNet might perform sub-optimally with one image while other DrosoNets actually output a correct prediction. Exploiting both of these features and the image-sequential nature of the SLAM environment, we propose a voting mechanism that takes the outputs of multiple trained DrosoNets and combines them to perform more reliable VPR. While this results in a larger and more complex algorithm, its inference time of 30ms and memory size of 12MBs is still vastly inferior to many state-of-the-art approaches such as VGG-16. Moreover, the developed voting system is not exclusive to DrosoNet as a baseline and can be utilized with any classifier-type algorithm that works on sequential imagery. Ideally, the baseline model should be compact, as multiple will be used, and present some degree of variation in its predictions for the voting process to take advantage of.

The remainder of this paper is structured as follows. Section \ref{relatedwork} gives an overview of related work in the VPR field. DrosoNet and the proposed voting algorithm are presented in detail in section \ref{method}. The experimental setup and evaluation criteria are explained in section \ref{exp}. Results are displayed and discussed in section \ref{results}. Finally, conclusions are drawn in Section \ref{conclusions}.

\section{Related Work}
\label{relatedwork}

Many approaches have been explored as possible solutions to the VPR problem. SURF \cite{ref:surf} and SIFT \cite{ref:surf} are examples of handcrafted feature-based models that employ local features for image matching and have been widely used for VPR. To complement these feature extractors, image retrieval algorithms such as Bag-of-words (BoW) have gotten a lot of attention as the process of searching the previously seen images for a match must be efficient to be performed in real time. Targeting robotics with constricted hardware, research has led to more compact BoW alternatives, employing binary BoW representations. Light models such as these are essential for robotic units that are equipped with low-end hardware but still require loop-closure processes for their SLAM navigation. While multiple local-feature based methods have been successful, such approaches face important challenges. The identification and extraction of descriptive and repeatable features in an image is an incredibly complicated task. Furthermore, these models usually do not face off well against large appearance changes that are bound to occur with long term robotic operations.

CNN based models have recently been introduced to the VPR field, achieving impressive performance results \cite{ref:VPR_CNNs}. These techniques rely on large amounts of annotated data to train deep neural networks and either use these models as whole-image descriptors or to collect local descriptors from the inner layers of the net. While the great performance of CNNs is indisputable, there are obvious shortcomings to this approach. In order to increase prediction accuracy, these networks have become deeper and more complex \cite{ref:mobilenets}, resulting in high memory and computational power needed to run these algorithms online, making them unsuitable for platforms equipped with low-end hardware, which is often the case in mobile robotic applications \cite{ref:res_hardware_1}. Recently, the interest in low-overhead CNN based algorithms has led to the development of more compact techniques. MobileNets \cite{ref:mobilenets} uses depth-wise separable convolutions to decrease model size and inference time and while it is primarily designed for image classification, it can be used for landmark recognition. Starting from a baseline AlexNet \cite{ref:alexnet}, SqueezeNet \cite{ref:squeezenet} achieves a much more compact model size by reducing filter sizes and down sampling in the deeper layers. By reducing a model's parameters precision after training, a processed named dynamic quantization, researchers were able to reduce memory requirements without significantly affecting performance \cite{ref:quant}.

Biological inspired algorithms are yet another approach to deal with the VPR problem in highly restricted platforms. Small animals are able to perform complex navigation tasks, such as localization, with neural activations \cite{ref:honeybee} \cite{ref:ants} which are simple and elegant when compared to artificial deep neural networks. Motivated by this observation, bio-models have been developed for general navigation \cite{ref:ratslam} and VPR \cite{ref:ogflynet}. Of particular interest for this work is the research conducted to understand the fruit fly's ability to navigate \cite{ref:droso_vpr} and process odors \cite{ref:ogflynet} by encoding complex patterns into compact representations, which inspired efficient lightweight algorithms for the VPR problem \cite{ref:flynetcann} and our proposed DrosoNet.

\section{Methodology}
\label{method}

We present two novel lightweight VPR algorithms as our contributions in this work: DrosoNet and the voting mechanism that builds on top of it. DrosoNet works by computing a low-memory representation of a given image. It is then trained as a classifier, where each place is a different class, that recognizes these small image tags and associates them with their respective place. The voting method exploits the stochastic nature of the DrosoNet training process, making use of several DrosoNet individuals to perform more accurate and consistent VPR while remaining compact relative to state-of-the-art approaches. The remaining of this section will explore both DrosoNet and the voting algorithm in depth.

\subsection{DrosoNet}
\label{drosonetexp}

\begin{figure*}[thpb]
\centering
\includegraphics[width=1\textwidth]{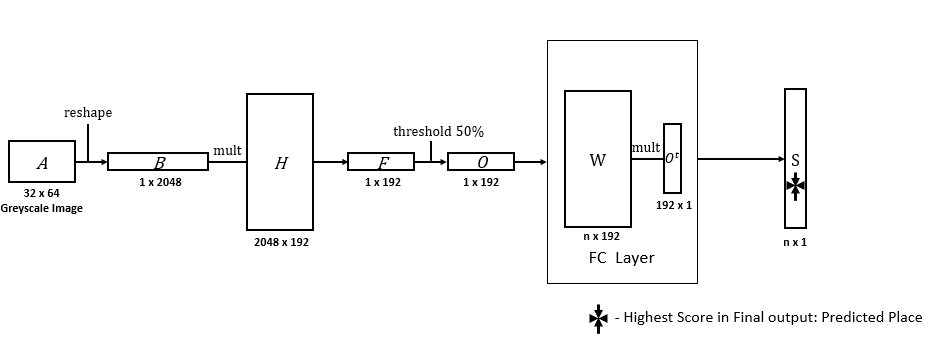}
\caption{DrosoNet implementation diagram. This process is repeated for each input image (each image corresponds to a place, n places). A is the input image; B is the input image reshaped to a row vector; B is a sparse binary matrix where each column has only 10\% of its values set to 1, the remaining to 0; F contains the 192 activation values; O is a binary image representation resulting from the threshold of the values of F; W is the weight matrix of the fully-connected classifier; S is the final output of the DrosoNet, where each value corresponds to a place score. After training takes place, DrosoNet is dynamically quantized to 8-bit precision, further reducing its memory usage.}
\label{DrosoNetdiagram}
\end{figure*}

DrosoNet is a bio-inspired model that draws inspiration from the fruit fly's brain circuits. The brain of these small insects is extremely efficient at recognizing different odors, especially when considering its size. While the VPR problem deals with visual information, the algorithm attempts to use a simplified version of the information processing that the fruit fly's brain uses for odor recognition.

The 
The fruit fly's neural circuits elaborate the information captured by olfactory sensors in three steps \cite{ref:fruitfly}. Firstly, a sort of normalization occurs, centering the mean of the activation rates of the flies' neurons for all odors. Secondly, around 10\% of the neurons that respond to an odor are evaluated and their activation rate is summed up. Finally, 5\% of the summed up values are used to create a binary representation of the given odor - this compact representation is then used to compare and recognize odors.

The proposed DrosoNet algorithm makes use of the fly's schema to encode a compact image representation that is then fed to a fully connected layer for classification. A simplified version of the process is designed and optimized for machine learning, Fig. \ref{DrosoNetdiagram} shows the operations that occur in the DrosoNet algorithm, displaying the dimensions of each matrix. The image is stretched into a row vector of size $1\times2048$. It is then multiplied by the matrix H, H is binary and sparse, with 10\% of the elements of each column randomly set to 1 and the remaining being 0 on the DrosoNet instantiating. This results in 10\% random pixels of the input image being taken into account when calculating the activation values, stored in F. The number of columns in H corresponds to the number of activations used, we set this value to 192. The top 50\% higher values in F are then set to 1 while the lower 50\% are set to 0, resulting in a binary representation for the input image, matrix O. O is then fed into a fully connected layer where the learning process takes place. The fully connected layer works as a classifier to predict the current place from the the vector O, hence including exactly one neuron per map's location. The highest value among the $n$ elements of the output vector is regarded as the matching location. Finally, after the model is trained, we reduce the parameters' precision down to 8-bit integers in a process named dynamic quantization, further reducing the memory size of DrosoNet.

\subsection{Voting Mechanism}
\label{swarmnet exp}

Our proposed voting mechanism combines the output of several DrosoNets to perform more effective and consistent VPR. In practice, the random initialization of DrosoNet's H matrix as well as the stochastic nature of training the fully-connected layer means that one particular DrosoNet might have a poor prediction for a given place while most other DrosoNets actually output an acceptable prediction. The voting exploits this observation and does not rely on any single DrosoNet to cast its prediction. Instead, it selects the prediction that most models agree on, following some specified ruling. The remaining of this section expands on our proposed voting mechanism.

We start by training a number of DrosoNets on the same training dataset and storing these trained models in a collection. We then run these models with the test dataset, obtaining the score vector and hence the predicted place for each DrosoNet. 

We now detail how the voting model works with $n$ trained DrosoNets to perform a prediction. The voting method takes into account the highest score given by the individual (the predicted place) as well as all the scores within a range around that predicted index. The selection of a score for a single DrosoNet, $d$, can be represented as
\begin{gather}
  v_i^d =
  \begin{cases}
                                   s_i^d & \text{if $l^d \leq i \leq u^d$} \\
                                   0 & \text{else} 
  \end{cases}
\end{gather}
where $s^d$ is the complete vector score given by the $d^{ith}$ DrosoNet after being normalized by a soft max function pass, $s_i^d$ denotes the $i^{th}$ score in $s^d$, $v_i^d$ is the $i^{th}$ score that is either copied from $s^d$ or set to $0$ and stored in vector $v^d$. $l^d$ and $u^d$ denote for the lower and upper index bounds of the scores to be selected around the highest score $d$, for the $d^{ith}$ DrosoNet, and are defined as
\begin{gather}
     p^d = argmax(s^d) \\
     l^d = max(0, p^d-r) \\
     u^d = min(len(s^d) - 1, p^d+r)
\end{gather}
where $p^d$ is the index of the highest score (hence the place predicted by the $d^{ith}$ DrosoNet), $r$ is a hyperparameter for the chosen range of selection and $len(s)$ is the length of the score vector $s^d$ (also corresponding to the number of places and hence is constant for all $n$ DrosoNets). The $max$ and $min$ functions are used to avoid negative and out-of-bound indexing. 

Using the above definitions, we construct the vector $v^d$ for each of the $n$ DrosoNets in the ensemble, where each score is set to either $0$ or the corresponding score in $s^d$. We obtain a single vector score $f$ by summing element-wise over all $n$ vectors $v^d$
\begin{gather}
    f = \sum_{d=1}^{n} v^d
\end{gather}
Finally, the index $m$ of the highest score in $f$ is selected as the matching place for the input image:
\begin{gather}
    m = argmax(f)
\end{gather}

\begin{figure*}[thpb]
\vspace*{1ex}
\centering
\includegraphics[width=1\textwidth]{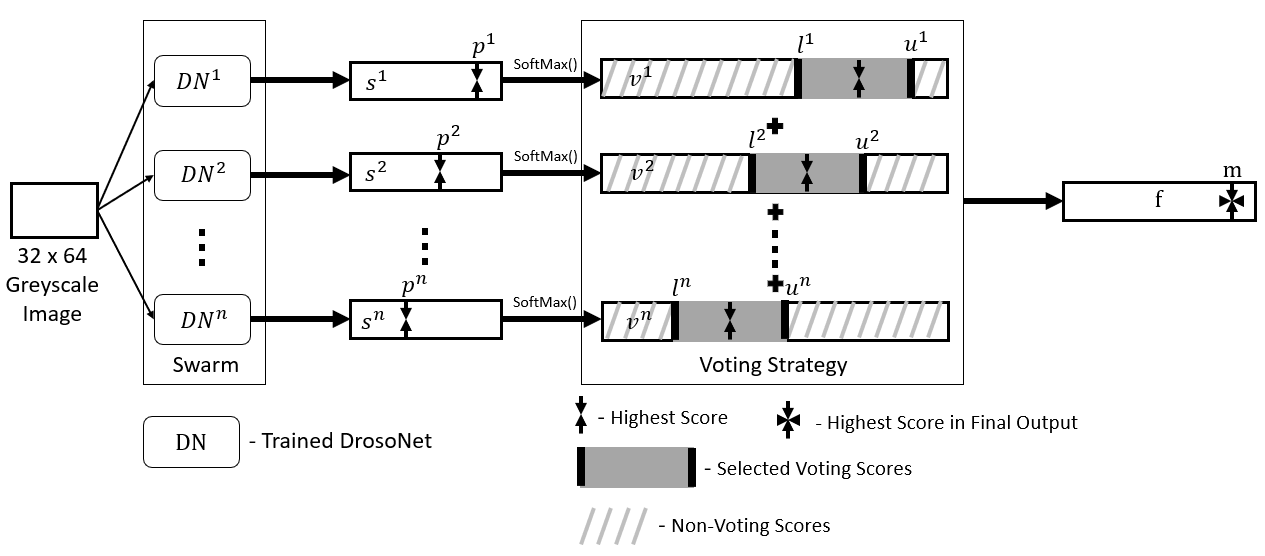}
\caption{Voting mechanism diagram, displaying the combination of several DrosoNets' outputs using our proposed voting method}
\label{swarmnetdiagram}
\end{figure*}

\section{Experiments}
We ran experiments with our proposed models as well as with FlyNet \cite{ref:flynetcann}, a published algorithm also inspired by the fruit fly's brain. We use different datasets to assess the models' capacity to deal with different VPR challenges: moderate to extreme appearance changes and small to moderate point-of-view (POV) variations. We note the performance of these algorithms as well as their memory usage and time required to process a single image. The remaining of this section provides details on models' settings, datasets and evaluation metrics.

\label{exp}
\subsection{Model Settings}

\begin{figure}[thpb]
\centering
\includegraphics[width=0.45\textwidth]{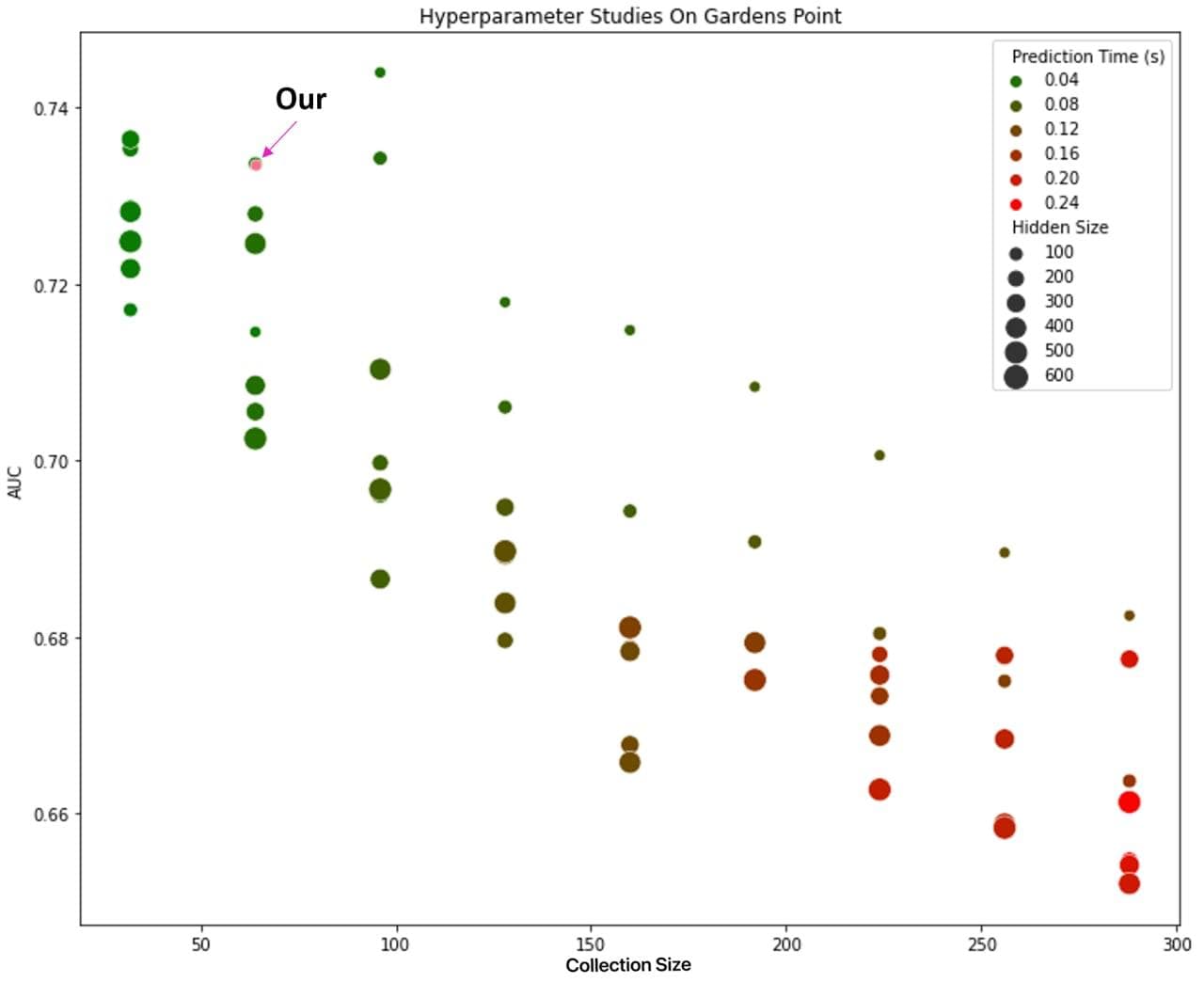}
\caption{Ablation studies results to determine the range of optimal hyperparameters for DrosoNet plus voting. We observe that a large collection of DrosoNets results in poorer VPR performance and increased computational cost, reducing the range of optimal parameters to a smaller number of DrosoNets.} 
\label{hyper_studies}
\end{figure}

For FlyNet, we use the exact same model settings and architecture as described in \cite{ref:flynetcann}.

For the standalone DrosoNet and for DrosoNet in conjunction with the voting mechanism, we conducted a series of ablation studies to find reasonable choices for the number of activations for DrosoNet and the number of models to be used in conjunction with the voting system. Fig. \ref{hyper_studies} presents the results of one of these experiments, clearly showing that, for the Gardens Point dataset, sticking to a smaller number of DrosoNets with a moderate number of activation yields the best AUC and efficiency results. By conducting such studies for the different datasets, we picked 192 as the number of neurons for DrosoNet and we use 64 DrosoNets for the voting collection as these values consistently showed strong and consistent results across all datasets. Also by experimentation, we select a value for the voting range $r$ equal to 50\% of the total number of places in the dataset, i.e. $r = 500$ for a dataset of 1000 images.

\subsection{Datasets}
\subsubsection{Nordland}
The Nordland dataset \cite{ref:nord} presents four distinct traversals of a train journey, one per season: Summer, Spring, Fall and Winter. This dataset is used to assess how a model deals with moderate to severe appearance changes. In our experiments, we utilize 1000 images per traversal, with models being trained on the Summer dataset and tested on the Fall and Winter sets, respectively assessing moderate and extreme appearance changes. A match is considered correct if the predicted place falls within 3 frame of the ground-truth image. Thus, for query image q and ground-truth image t, images t-1 to t+1 would be considered correct matches.

\subsubsection{Gardens Point}
The Gardens Point dataset is recorded in the Queensland University of Technology. We utilize two distinct traversals of 200 images from this dataset to assess model performance on strong point of view variations. Both traversals were captured during the day, with the second being laterally shifted to the right. A match is considered correct if the predicted place falls within 5 frames of the ground-truth image. Thus, for query image q and ground-truth image t, images t-2 to t+2 would be considered correct matches.

\subsubsection{Oxford RobotCar}
The Oxford RobotCar \cite{ref:robotoxford} traversals used present challenging illumination changes. We utilize two traversals of 200 images each in our experiments. We allow for a 10 frame margin tolerance around the ground truth location, as per previous research \cite{ref:prevres1}, \cite{ref:prevres2}. Thus, for query image q and ground-truth image t, images t-10 to t+10 would be considered correct matches.

\subsubsection{Lagout 15 Degrees POV Variation} 
The Lagout 15 synthetic dataset consists of aerial footage captured at a 15 degree angle. It assesses model performance on moderate POV variations with 6 degrees-of-freedom movement. We train the 3 models on the Lagou traversal at a 0 degree angle and test on Lagout 15, utilizing 200 images per traversal and allowing for 3 frame margin of error around the ground-truth location frame.

\subsubsection{Corvin 30 Degrees POV And Scale Variation}
The Corvin 30 synthetic dataset was recorded using flight imagery of the Corvin Castle at a 30 degrees angle. It is intended to assess model resilience on strong point-of-view and zoom variations when allowing 6 DOF movements. We train our models using the Corvin traversal at a 0 degrees angle and then test on Corvin 30, utilizing 1000 images per traversal and allowing for a 20 frame margin of error around the ground-truth location frame.

\subsection{Evaluation Metrics}
Precision-Recall (PR) curves are often utilized to evaluate the performance of VPR techniques \cite{ref:pr_jus1}, \cite{ref:pr_jus2} as they are preferable when dealing with imbalanced datasets. We utilize PR curves and the Area Under these curves to assess the peformance of the different models.

Furthermore, we are interested in the size and complexity of each model as we are focusing on developing extremely compact algorithms. To evaluate compactness, we show the memory usages of the proposed models and common state-of-the-art algorithms. 

\section{Results and Analysis}
In this section, we interpret the results obtained by experiments in both fronts: performance and memory usage of the tested models. We also take a look at these metrics for popular state-of-the-art approaches and highlight the compromise between performance and compactness.

\label{results}
\subsection{VPR Performance}
This sub-section outlines the PR-curves and AUC under these curves for the 3 tested VPR techniques for the described benchmark datasets.

\subsubsection{Nordland Fall}
\begin{figure}[thpb]
\vspace*{1ex}
\centering
\includegraphics[width=0.45\textwidth]{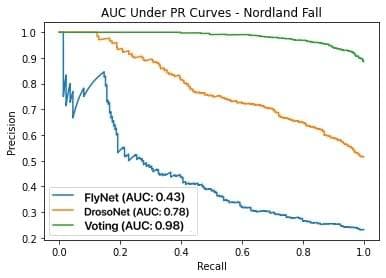}
\caption{AUC under PR curves for the tested models on the Nordland Fall dataset, assessing moderate appearance changes. Both standalone DrosoNet and the voting approach outperform FlyNet. Notably, DrosoNet based voting achieves an AUC of 0.98, being close to a perfect classifier.}
\label{pr_fall}
\end{figure}

For the Nordland Fall dataset, Fig. \ref{pr_fall} shows the PR-curves obtained by the tested models as well as the AUC values under these curves. This dataset presents moderate appearance changes, with DrosoNet outperforming FlyNet and the voting strategy in turn severely outperforming both of these models. In fact, DrosoNet plus voting achieves an AUC value of 0.98, being close to a perfect classifier. All of these 3 techniques act as whole-image descriptors and we expected them to achieve their best performance with appearance variations.

\subsubsection{Nordland Winter} 
\begin{figure}[thpb]
\centering
\includegraphics[width=0.45\textwidth]{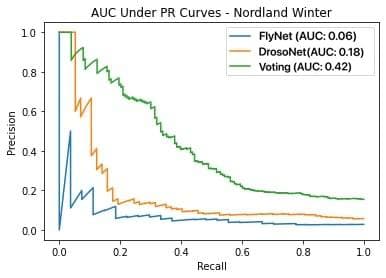}
\caption{AUC under PR curves for the tested models on the Nordland Winter dataset, assessing extreme appearance changes. As expected, all 3 models perform worse compared to the Fall testing. The voting strategy continues to outperform the other two models and manages to retain competitive results.}
\label{pr_winter}
\end{figure}

PR-curves and respective AUCs for the Nordland Winter dataset are illustrated in Fig. \ref{pr_winter}. This dataset contains extreme appearance changes, as such being much more challenging than the previous Nordland Fall dataset. While DrosoNet continues to outperform FlyNet, neither of these techniques achieves a competitive AUC value. The DrosoNet plus voting approach also sees a performance drop compared to Fall but manages to maintain respectable performance at an AUC value of 0.42. 


\subsubsection{Gardens Point}
\begin{figure}[thpb]
\vspace*{1ex}
\centering
\includegraphics[width=0.45\textwidth]{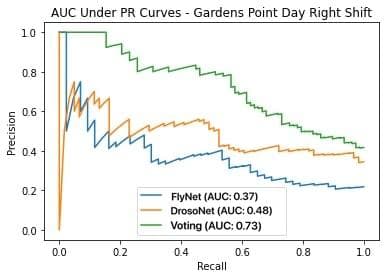}
\caption{AUC under PR curves for the tested models on the Garden Points day right shift dataset, assessing strong viewpoint variation.}
\label{pr_right}
\end{figure}

Fig. \ref{pr_right} shows the PR-curves and AUCs for the Gardens Point right shift dataset. This dataset tests how the VPR techniques perform with a moderate, lateral view-point change. We continue to observe that standalone DrosoNet achieves higher performance than FlyNet, with DrosoNet plus voting being the best performing model of the 3. Being whole-image techniques, we expect all 3 models to struggle with POV variation challenges. Indeed, even for a moderate, unilateral POV change, we see a performance decrease compared to the results obtained with a moderate appearance change in Nordland Fall. Nevertheless, the use of multiple models for voting continues to show promise, achieving a competitive AUC of 0.73.

\subsubsection{Oxford RobotCar}
\begin{figure}[thpb]
\vspace*{1ex}
\centering
\includegraphics[width=0.45\textwidth]{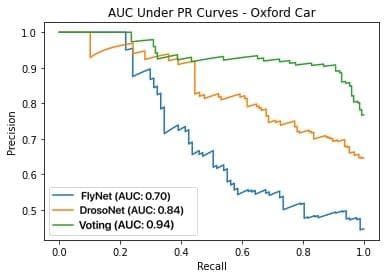}
\caption{AUC under PR curves for the tested models on the Oxford Car dataset, assessing moderate viewpoint and illumination changes.}
\label{pr_ox}
\end{figure}

This dataset presents mostly illumination changes and model's results can be found in Fig. \ref{pr_ox}. As illumination variation is closely related to appearance change, we expected the models to perform well with this testing. We observe that all the 3 models perform well and the voting mechanism continues to show better results over DrosoNet and FlyNet.

\subsubsection{Lagout 15 Degrees POV Variation}
\begin{figure}[thpb]
\vspace*{1ex}
\centering
\includegraphics[width=0.45\textwidth]{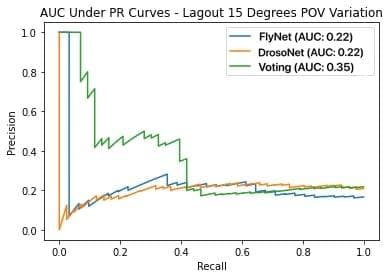}
 \cite{}\caption{AUC under PR curves for the tested models on the Lagout 15 degrees variation dataset, assessing moderate 6 DOF viewpoint changes.}
\label{pr_lag15}
\end{figure}

Model's performance on this dataset can be observed in Fig. \ref{pr_lag15}. Lagout 15 presents moderate POV variations with 6 degrees-of-freedom, being particularly challenging for whole-image techniques as is the case for the 3 tested models. Accordingly, we observe that neither of the models achieves competitive AUC values. Despite the overall poor performance, DrosoNet based voting continues outperforming both standalone DrosoNet and FlyNet, showing that the use of combination of multiple DrosoNets achieves higher performance than reliance on a single one.

\subsubsection{Corvin 30 Degrees POV And Scale Variation}
\begin{figure}[thpb]
\vspace*{1ex}
\centering
\includegraphics[width=0.45\textwidth]{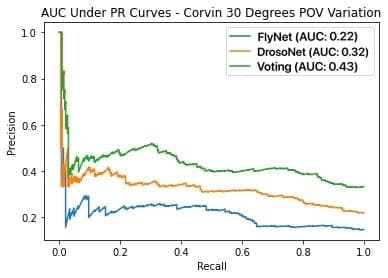}
\caption{AUC under PR curves for the tested models on the Corvin 30 degrees and scale variation dataset, assessing strong viewpoint and scale changes under 6 DOF movement.}
\label{pr_corvin30}
\end{figure}

This dataset exhibits challenging POV and zoom variations, characteristic of 6 DOF movement, the obtained PR-curves and respective AUCs can be seen in Fig. \ref{pr_corvin30}. We expected this type of variations to be the most challenging to the tested models and indeed, none of the models seems adequate having in account the large 20 frame margin for error. Nevertheless, we observe that the voting approach is once again the highest performing model, indicating that the use of multiple individuals is still superior in the 6 degrees-of-freedom POV and zoom variation case.

\subsection{Computational Resources}

As mentioned, the target applications for our proposed models are robotic applications with extremely constricted hardware and as such, computational resources are an important aspect of our algorithms. Computational complexity is usually the main bottleneck on heavily restricted hardware platforms, GPUs being often needed to efficiently compute convolution operations. In this section, we focus on two evaluation metrics, inference times with respective frames-per-second rates and memory size, showing how the different tested models compare in relation to their computational efficiency. For further comparison, we also include the same metrics for the VGG16 model, tested under the same hardware conditions. 

Table \ref{memory&time} shows exactly the aforementioned metrics, obtained while running the different models in a Ryzen 7 4000 Series processor. Our choice to utilize a CPU rather than a GPU is motivated by the fact that CPUs are available even in the lowest-end of hardware, when GPUs are usually not present in extremely constrained mobile robots. 

\begin{table}[htbp]
  \centering
  \caption{Inference times and memory usage comparison}
    \begin{tabular}{lrrr}
          & \multicolumn{1}{l}{Prediction Time (ms)} & \multicolumn{1}{l}{FPS} &
          \multicolumn{1}{l}{Memory Size(MB)}\\
    FlyNet & 1 & 1000 & 0.26\\
    DrosoNet & 1 & 1000 & 0.19\\
    Voting & 30  & 33 & 12.16 \\
    VGG16 & 200   & 5 & 526\\
    \end{tabular}%
  \label{memory&time}%
\end{table}%

The inference time for a single image is the same for both FlyNet and a single DrosoNet, at 1ms both of these algorithms are remarkably fast, efficient and capable of processing upwards of 1000 frames per second on the tested hardware. While 1000FPS is an impressive rate, a platform would need to be equipped with a high-frame-rate enabled camera to make full use of these FPS values. The voting mechanism on multiple DrosoNets shows a prediction time of 30ms, a considerable increase compared to the previously mentioned models achieving a 33FPS rate. While it is a considerate drop in the important FPS metric, 30FPS is a standard recording setting for common cameras which the voting method can take advantage of without the need for more powerful sensors. Despite this expected increase in computational complexity, utilizing the voting system remains around 7 times faster than VGG16 which inference time is of 200ms and respective FPS rate of 5.

In terms of memory size, DrosoNet is the smallest model at 190KiBs. It is even more compact than FlyNet even though it contains more parameters, due to the 8-bit quantization employed. As expected, utilizing a collection of DrosoNets for voting is less compact than both FlyNet and a single DrosoNet,  for a total model size of 12.16MBs. Nevertheless, it remains extremely compact when to compared to the popular VGG16 model which comes at 526MBs. While memory size is usually not a issue for most practical applications, we emphasize just how small our proposed models are, being suitable for even the smallest of platforms.

\section{Conclusions and Future Work}
In this work, two techniques are proposed to address the nneed for lightweight VPR algorithms for the most hardware restricted of robotic platforms. We first introduce DrosoNet, an extremely compact algorithm inspired by the brain of the fruit fly. Relative to its size, it obtains respectable results in the benchmark datasets, especially when dealing with moderate appearance changes. However, most systems require more robust VPR than what DrosoNet is able to achieve. Our solution is to employ a voting scheme across multiple DrosoNets, exploiting the low memory usage and variation of DrosoNet, utilizing multiple of these small models to achieve competitive VPR performance while remaining compact relatively to CNN based algorithms. When comparing the trade-off between size and performance, the DrosoNet based voting model stands as a compact VPR algorithm with competitive performance, suitable for hardware-restrictive robotic applications.

For further research, one should focus on how to improve the performance of standalone DrosoNet and DrosoNet coupled with voting for extreme POV variation challenges, as the models currently struggle in these settings. Furthermore, investigating how different DrosoNets, or any other baseline voting classifier, complement each other can prove beneficial to select only the most informative models. Finally, testing the proposed voting mechanism with different suitable baseline models is another viable route for continuation in this research.

\label{conclusions}

\addtolength{\textheight}{-12cm}   




\bibliographystyle{IEEEtran}
\bibliography{ref}

\begin{thebibliography}{10}
\providecommand{\url}[1]{#1}
\csname url@rmstyle\endcsname
\providecommand{\newblock}{\relax}
\providecommand{\bibinfo}[2]{#2}
\providecommand\BIBentrySTDinterwordspacing{\spaceskip=0pt\relax}
\providecommand\BIBentryALTinterwordstretchfactor{4}
\providecommand\BIBentryALTinterwordspacing{\spaceskip=\fontdimen2\font plus
\BIBentryALTinterwordstretchfactor\fontdimen3\font minus
  \fontdimen4\font\relax}
\providecommand\BIBforeignlanguage[2]{{%
\expandafter\ifx\csname l@#1\endcsname\relax
\typeout{** WARNING: IEEEtran.bst: No hyphenation pattern has been}%
\typeout{** loaded for the language `#1'. Using the pattern for}%
\typeout{** the default language instead.}%
\else
\language=\csname l@#1\endcsname
\fi
#2}}

\bibitem{ref:vpr-survey}
S.~Lowry and et~al., ``Visual place recognition: A survey,'' \emph{IEEE T-RO,
  vol. 32, no. 1, pp. 1–19}, 2016.

\bibitem{ref:season_changes}
T.~Naseer, L.~Spinello, W.~Burgard, and C.~Stachniss, ``Robust visual robot
  localization across seasons using network flows,'' \emph{Twenty Eighth AAAI
  Conference on Artificial Intelligence}, 2014.

\bibitem{ref:pov_changes}
A.~Pronobis, B.~Caputo, P.~Jensfelt, and H.~I. Christensen, ``A discriminative
  approach to robust visual place recognition,'' \emph{IEEE}, vol. IROS, pp.
  3829--3836, 2006.

\bibitem{ref:illu_changes}
A.~Ranganathan, S.~Matsumoto, and D.~Ilstrup, ``Towards illumination invariance
  for visual localization,'' \emph{IEEE}, vol. ICRA, pp. 3791--3798, 2013.

\bibitem{ref:dyna_changes}
C.-C. Wang and et~al., ``Simultaneous localization, mapping and moving object
  tracking,'' \emph{IJRR}, vol.~26, pp. 889--916, 2017.

\bibitem{ref:VPR_CNNs}
Z.~C. et~al., ``Convolutional neural network-based place recognition,''
  \emph{Proc. Australas. Conf. Robot. Autom.}, 2014.

\bibitem{ref:res_hardware_1}
F.~Maffra, Z.~Chen, and M.~Chli, ``Viewpoint-tolerant place recognition
  combining 2d and 3d information for uav navigation,'' \emph{IEEE}, vol. 2018
  IEEE International Conference on Robotics and Automation (ICRA), pp.
  2542--2549, 2018.

\bibitem{ref:res_hardware_2}
B.~Ferrarini, M.~Waheed, S.~Waheed, S.~Ehsan, M.~Milford, and K.~D.
  McDonald-Maier, ``Visual place recognition for aerial robotics: Exploring
  accuracy-computation trade-off for local image descriptors,'' \emph{IEEE},
  vol. 2019 NASA/ESA Conference on Adaptive Hardware and Systems (AHS), pp.
  103--108, 2019.

\bibitem{ref:flynetcann}
M.~C. et~al., ``A hybrid compact neural architecture for visual place
  recognition,'' \emph{IEEE ROBOTICS AND AUTOMATION LETTERS. PREPRINT VERSION},
  Dec. 2019.

\bibitem{ref:squeezenet}
\BIBentryALTinterwordspacing
F.~N. Iandola, M.~W. Moskewicz, K.~Ashraf, S.~Han, W.~J. Dally, and K.~Keutzer,
  ``Squeezenet: Alexnet-level accuracy with 50x fewer parameters and
  {\textless}1mb model size,'' \emph{CoRR}, vol. abs/1602.07360, 2016.
  [Online]. Available: \url{http://arxiv.org/abs/1602.07360}
\BIBentrySTDinterwordspacing

\bibitem{ref:mobilenets}
\BIBentryALTinterwordspacing
A.~G. Howard, M.~Zhu, B.~Chen, D.~Kalenichenko, W.~Wang, T.~Weyand,
  M.~Andreetto, and H.~Adam, ``Mobilenets: Efficient convolutional neural
  networks for mobile vision applications,'' \emph{CoRR}, vol. abs/1704.04861,
  2017. [Online]. Available: \url{http://arxiv.org/abs/1704.04861}
\BIBentrySTDinterwordspacing

\bibitem{ref:fruitfly}
A.~G. C.~Pehlevan and D.~B. Chklovskii, ``A clustering neural network model of
  insect olfaction,'' \emph{Asilomar Conf. Signals, Systems, and Computers},
  vol.~51, p. 593–600, 2017.

\bibitem{ref:surf}
H.~Bay, A.~Ess, T.~Tuytelaars, and L.~V. Gool, ``Speeded-up robust features
  (surf),'' \emph{Computer vision and image understanding, vol. 110, pp.
  346-359}, 2008.

\bibitem{ref:alexnet}
A.~Krizhevsky, I.~Sutskever, and G.~E. Hinton, ``Imagenet classification with
  deep {Convolutional Neural Networks},'' in \emph{Advances in neural
  information processing systems}, 2012, pp. 1097--1105.

\bibitem{ref:quant}
Y.~Xu, S.~Zhang, Y.~Qi, J.~Guo, W.~Lin, and H.~Xiong, ``Dnq: Dynamic network
  quantization,'' 2018.

\bibitem{ref:honeybee}
A.~C. et~al, ``The green brain project—developing a neuromimetic robotic
  honeybee,'' \emph{Biom. and Biohybrid Syst}, pp. 362--363, 2013.

\bibitem{ref:ants}
A.~N. et~al., ``Mapping the navigational knowledge of individually foraging
  ants, myrmecia croslandi,'' \emph{Proc. R. Soc. B.}, vol. 280, 2013.

\bibitem{ref:ratslam}
M.~J.~M. et~al, ``Ratslam: a hippocampal model for simultaneous localization
  and mapping,'' \emph{Proc. IEEE Int. Conf. Robot. Autom}, vol.~1, p.
  403–408, 2004.

\bibitem{ref:ogflynet}
S.~D. et~al, ``A neural algorithm for a fundamental computing problem,''
  \emph{Science}, vol. 358, p. 793–796, 2017.

\bibitem{ref:droso_vpr}
T.~A. Ofstad, C.~S. Zuker, and M.~B. Reiser, ``Visual place learning in
  drosophila melanogaster,'' \emph{Nature}, vol. 474, pp. 362--363, 2011.

\bibitem{ref:nord}
P.~N. N.~S~underhauf and P.~Protzel, ``Are we there yet? challenging seqslam on
  a 3000 km journey across all four seasons,'' \emph{Proc. Workshop Long-Term
  Autonomy IEEE Int. Conf. Robot. Autom}, 2013.

\bibitem{ref:robotoxford}
k.~T. O. R.~D. 1~Year, ``W. maddern et al,'' \emph{Int. J. Robot. Res},
  vol.~36, pp. 3--15, 2017.

\bibitem{ref:prevres1}
N.~S. S.~Garg and M.~Milford, ``Lost? appearance-invariant place recognition
  for opposite viewpoints using visual semantics,'' \emph{Proc. Robot.: Sci.
  Syst}, 2018.

\bibitem{ref:prevres2}
A.~J. S.~Hausler and M.~Milford, ``Multi-process fusion: Visual place
  recognition using multiple image processing methods,'' \emph{IEEE Robot.
  Autom. Lett}, vol.~4, p. 1924–1931, 2019.

\bibitem{ref:pr_jus1}
J.~Davis and M.~Goadrich, ``The relationship between precision-recall and roc
  curves,'' \emph{Proc. 23rd ACM Int. Conf. Mach. Learn.}, pp. 233--240, 2006.

\bibitem{ref:pr_jus2}
D.~M. Powers, ``Evaluation: From precision, recall and f-measure to roc,
  informedness, markedness and correlation,'' \emph{J. Machine Learn. Tech.},
  vol.~2, pp. 37--63, 2011.

\end{thebibliography}

\end{document}